\documentclass[9pt,conference]{IEEEtran}
\IEEEoverridecommandlockouts
\usepackage{url}
\usepackage{booktabs}
\usepackage{amsfonts}
\usepackage{nicefrac}
\usepackage{microtype}
\usepackage{graphicx}

\usepackage[caption=false,font=footnotesize]{subfig}

\usepackage{cite}
\usepackage{mathtools}
\ifCLASSOPTIONcompsoc
  \usepackage[caption=false,font=normalsize,labelfont=sf,textfont=sf]{subfig}
\else
  \usepackage[caption=false,font=footnotesize]{subfig}
\fi
\usepackage{pgfplots}
\pgfplotsset{compat=1.10}
\usetikzlibrary{arrows,positioning,automata,bayesnet}

\begin{document}

\title{Structured Sparse Modelling with Hierarchical GP}

\author{\IEEEauthorblockN{Danil Kuzin, Olga Isupova, Lyudmila Mihaylova}
\IEEEauthorblockA{The University of Sheffield, Sheffield, UK\\
Email: dkuzin1@sheffield.ac.uk, oisupova1@sheffield.ac.uk, l.s.mihaylova@sheffield.ac.uk}%
\thanks{Olga Isupova and Lyudmila Mihaylova acknowledge the support from the EC Seventh Framework Programme [FP7 2013-2017] TRAcking in compleX sensor systems (TRAX) Grant agreement no.: 607400.}
}

\maketitle

\begin{abstract}
In this paper a new Bayesian model for sparse linear regression with a spatio-temporal structure is proposed. It incorporates the structural assumptions based on a hierarchical Gaussian process prior for spike and slab coefficients. We design an inference algorithm based on Expectation Propagation and evaluate the model over the real data.
\end{abstract}

\IEEEpeerreviewmaketitle

\section{Introduction}

Sparse regression problems arise often in various applications, e.g., model selection, compressive sensing, EEG source localisation and gene modelling~\cite{Bach14,hastie2015statistical}. One of the Bayesian approaches to force the coefficients being zeros is the spike and slab prior~\cite{MitchellBeauchamp1988}: each component is modelled as a mixture of spike, that is the delta-function in zero, and slab, that is some vague distribution. Following the Bayesian approach, latent variables that are indicators of spikes are added to the model~\cite{PolsonScott2010} and the relevant distribution is placed over them~\cite{murphy2012machine}.

In this model each component is modelled to be spike or slab independently. However, in many applications non-zero elements tend to appear in groups forming an unknown structure: wavelet coefficients of images are usually organised in trees \cite{Mallat2008}, chromosomes have a spatial structure along the genome \cite{hastie2015statistical}. 

We propose an extension of the spike and slab model by imposing a hierarchical Gaussian process (GP) prior on the latent variables. Such hierarchical prior allows to model spatial structural dependencies for coefficients that can evolve in time. The new model is flexible as spatial and temporal dependencies are decoupled by different levels of the hierarchical GP prior. 

\section{Proposed model}
\label{sec:model}
The observations $\mathbf{y}_t \in \mathbb{R}^K$ are collected with the design matrix $\mathbf{X} \in \mathbb{R}^{K\times N}$ from the unknown coefficients $\boldsymbol{\beta}_t \in \mathbb{R}^N$ at every time moment $t\in [1, \ldots, T]$, with independent noise:
\begin{equation}
\mathbf{y}_t  \sim \mathcal{N}(\mathbf{y}_t | \mathbf{X}\boldsymbol{\beta}_t, \sigma^2 \mathbf{I}).
\end{equation}
We consider the case when $K < N$, therefore the problem of recovery of $\boldsymbol{\beta}_t$ from $\mathbf{y}_t$ is ill-posed and regularisation is required.  

\paragraph{Sparsity}
The vectors $\boldsymbol\beta_t$ are assumed to be sparse, that is implemented in the model using the spike and slab approach:
\begin{equation}
\beta_{it} \sim \omega_{it}\delta_0(\beta_{it})+(1-\omega_{it})\mathcal{N}(\beta_{it} | 0, \sigma_\beta^2),
\end{equation}
where $\omega_{it}$ are the latent indicators of spike and slab.

\paragraph{Spatial clustering}
Non-zero elements in $\boldsymbol{\beta}_t$ are assumed to be clustered in groups at every timestamp. Therefore spatial dependencies for the positions of spikes in $\beta_{it}$ are modelled with the GP:
\begin{align}
\omega_{it} &\sim \text{Ber}(\omega_{it} | \Phi(\gamma_{it})),\quad \Phi(\cdot) \text{ is the standard Gaussian cdf}\\
\boldsymbol\gamma_t &\sim \mathcal{N}(\boldsymbol\gamma_t | \boldsymbol\mu_t, \boldsymbol\Sigma_0),\quad \boldsymbol\Sigma_0(i,j) = \alpha_{\Sigma}\exp\left(-\frac{(i-j)^2}{2 l^2_{\Sigma}}\right).
\end{align}

GPs specify prior over an unknown structure. This is particularly useful as it allows to avoid a specification of any structural patterns --- structural modelling is governed only by the GP covariance function.

\paragraph{Temporal evolution}
Clusters of spikes in $\boldsymbol{\beta}_t$ are assumed to evolve in time. This evolution is addressed with the hierarchical GP dynamic system model~\cite{deisenroth2012expectation}. The mean for the spatial GP changes over time according to the top-level temporal GP:
\begin{equation}
\boldsymbol\mu_t \sim \mathcal{N} (\boldsymbol\mu_t | \boldsymbol\mu_{t-1}, \mathbf{W}), \quad \mathbf{W}(i,j) = \alpha_{W}\exp\left(-\dfrac{(i-j)^2}{2 l^2_W}\right).
\end{equation}
This allows to implicitly specify the prior over the transition function of the structure. The rate of the evolution is controlled with the top-level GP covariance function. 

The exact posterior of the parameters is intractable, therefore approximate inference methods are required. Inference is based on Expectation Propagation~\cite{minka2001expectation} in this paper.

The structural assumptions in sparse models are studied in the literature. The group lasso~\cite{yuan2006model} provides sparse solutions for predefined groups of coefficients. Group constraints for sparse models include smooth relevance vector machines~\cite{schmolck2008smooth}, Boltzmann machine prior~\cite{dremeau2012boltzmann};
spatio-temporal coupling of the parameters~\cite{van2010efficient, wu2014sparse}. In~\cite{andersen2015bayesian} a spatio-temporal structure is modelled with a one-level GP prior. In contrast to that model the new one introduces an additional level of a GP prior for temporal dependencies, therefore the temporal and spatial structures are decoupled, adding flexibility to the model. The high-level GP controls the change of spike groups in time while the low-level GP allows the local changes within each group.

\section{Numerical experiments}
\label{sec:experiments}
The performance of the proposed hierarchical GP algorithm is compared with the one-level GP prior introduced in~\cite{andersen2015bayesian}.
We apply both algorithms to the problem of object detection in video sequences. The Convoy dataset~\cite{Warnell2014} is used where the frame difference is applied for moving object detection. The sparse observations are obtained as $\mathbf{y}_t = \mathbf{X} \boldsymbol\beta_t$, where $\mathbf{X}$ is the matrix with i.i.d. Gaussian elements, the number of observations is $\approx 40\%$ of the dimension of the hidden signal $\boldsymbol\beta_t$. This procedure corresponds to compressive sensing observations~\cite{Cevher2008}. 

\begin{figure*}[!t]
\centering
\subfloat[Original frame]{\includegraphics[width=0.24\textwidth]{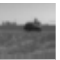}
\label{subfig:original}}
\hfil
\subfloat[Frame with subtracted background]{\includegraphics[width=0.24\textwidth]{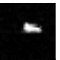}
\label{subfig:true_subtracted}}
\hfil
\subfloat[Hierarchical GP reconstruction]{\includegraphics[width=0.24\textwidth]{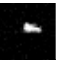}
\label{subfig:our_car}}
\hfil
\subfloat[One-level GP reconstruction]{\includegraphics[width=0.24\textwidth]{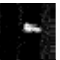}
\label{subfig:andersen_car}}
\caption{Sample frame with reconstruction results from sparse measurements. \protect\subref{subfig:original}: the original non-compressed frame; \protect\subref{subfig:true_subtracted}: object detection results based on non-compressed frame difference (static background frame is subtracted from the original frame); \protect\subref{subfig:our_car}: reconstruction of compressed object detection results based on the proposed hierarchical GPs method; \protect\subref{subfig:andersen_car}: reconstruction of the compressed object detection results based on the one-level GP method. \protect\subref{subfig:true_subtracted} represents the true $\boldsymbol\beta_t$, while \protect\subref{subfig:our_car} and \protect\subref{subfig:andersen_car} are posterior estimates $\widehat{\boldsymbol\beta}_t$ obtained by the reconstruction algorithms.}
\label{pic:reconstruction}
\end{figure*}

The reconstruction results on the sample frames are presented in Figure~\ref{pic:reconstruction}. The obtained performance measures values can be found in Table~\ref{tab:numerical_results}. The F-measure compares binary masks computed by thresholding the true vectors~$\boldsymbol\beta$ and the posterior estimates~$\widehat{\boldsymbol\beta}$. NMSE represents normalised mean squared error between $\boldsymbol\beta$ and $\widehat{\boldsymbol\beta}$. The proposed algorithm shows better results in terms of both measures. 

\section{Conclusion}

In this work we propose a new approach for spatial structure modelling in sparse models that allows to capture complex temporal evolution of data patterns. We also develop an efficient inference method based on EP. The numerical experiments show superiority of the proposed model over the current state-of-the-art. 

\begin{table}[t]
  \caption{Performance of signal reconstruction for background subtraction in video}
  \label{tab:numerical_results}
  \centering
  \begin{tabular}{lll}
    \toprule
    Measure     &  Hierarchical GP     & One-level GP \\
    \midrule
    NMSE & $\mathbf{0.0153}$  &   $0.0392$   \\
    F-measure & $\mathbf{0.9870}$ &  $0.9403$     \\
    \bottomrule
  \end{tabular}
\end{table}

\small
\bibliography{References}

\end{document}